\documentclass{article}

     \usepackage[nonatbib,preprint]{neurips_2020}

\usepackage[utf8]{inputenc} 
\usepackage[T1]{fontenc}    
\usepackage{hyperref}       
\usepackage{url}            
\usepackage{booktabs}       
\usepackage{amsfonts}       
\usepackage{nicefrac}       
\usepackage{microtype}      

\usepackage{graphicx}
\usepackage{caption}
\usepackage{subcaption}
\usepackage{wrapfig}

\usepackage[style=ieee, 
citestyle=numeric-comp,
sorting=none, backend=bibtex]{biblatex}
\title{Know Where To Drop Your Weights: Towards Faster Uncertainty Estimation}

\author{%
  Akshatha Kamath\\
  Department of Computer Science \& Engineering\\
  Manipal Institute of Technology\\
  576104 Karnataka, India \\
  \texttt{akshutk@gmail.com} \\
  \And
  Dwaraknath Gnaneshwar\\
  Department of Information Technology\\
  Manipal Institute of Technology\\
  576104 Karnataka, India\\
  \texttt{dwarakasharma@gmail.com}
  \And
  Matias Valdenegro-Toro\\
  German Research Center for Artificial Intelligence\\
  28359 Bremen, Germany\\
  \texttt{matias.valdenegro@dfki.de}
  
}

\begin{document}

\maketitle

\begin{abstract}
Estimating epistemic uncertainty of models used in low-latency applications and Out-Of-Distribution samples detection is a challenge due to the computationally demanding nature of uncertainty estimation techniques. Estimating model uncertainty using approximation techniques like Monte Carlo Dropout (MCD), DropConnect (MCDC) requires a large number of forward passes through the network, rendering them inapt for low-latency applications. 
We propose Select-DC which uses a subset of layers in a neural network to model epistemic uncertainty with MCDC. Through our experiments, we show a significant reduction in the GFLOPS required to model uncertainty, compared to Monte Carlo DropConnect, with marginal trade-off in performance. We perform a suite of experiments on CIFAR 10, CIFAR 100, and SVHN datasets with ResNet and VGG models. We further show how applying DropConnect to various layers in the network with different drop probabilities affects the networks performance and the entropy of the predictive distribution. 
\end{abstract}

\section{Introduction}
Deep neural networks are increasingly playing an important role in every industry. Their prowess as function approximators that are trainable using gradient-based optimization methods and as feature extractors for huge amounts of data has made them extremely successful in Computer Vision, Natural Language Processing, Reinforcement Learning, etc. Many products and in-production systems, which use Deep Learning as a back-end, deal with sensitive data and mission-critical subsystems. It is imperative that we are able to guarantee the safety of such systems and make them robust to faults \cite{goodfellow2014explaining}. Identifying instances where the network is uncertain about its follow through and quantifying model confidence is important to guarantee a fail-safe mechanism \cite{kendall2015bayesian}\cite{loquercio2020general}. Once we identify that a network is unsure about its predictions, the control can be transferred to a human-in-the-loop to take over.

Modern neural networks are poorly calibrated \cite{guo2017calibration}, i.e, the probability associated with the predicted class label is not a good representative of the true correctness likelihood. Bayesian Neural Networks \cite{denker1991transforming}, \cite{mackay1992practical}
(BNN) combine the strengths of neural networks and stochastic modelling. 
Networks are usually trained with maximum likelihood (MLE) or Maximum a Posteriori (MAP) estimation. 
Instead of the point estimates that we get from both MLE and MAP, we impose a full posterior distribution over the parameters of the network that takes weight uncertainty into account. Finding the posterior analytically, however, is computationally intractable. Therefore, we approximate the true posterior with a variational distribution $q(w|\theta)$ by minimizing the Kullback-Leibler(KL) Divergence between the variational distribution (such as a Gaussian) and the true distribution $p(w|D)$, where $D$ is the dataset.

Uncertainty estimation in neural networks is a compute-intensive process, even using simple approximations like Monte Carlo Dropout \cite{gal2016dropout} and Monte Carlo DropConnect(MCDC)\cite{mobiny2019dropconnect}. In this paper, we propose a modified MCDC, that has significant computational gains over a vanilla MCDC implementation.

In this work we explore how to reduce the computational requirements to make forward passes of a DropConnect enabled network, and observe some interesting results. Applying DropConnect to a network improves its accuracy compared to a baseline without DropConnect, and the magnitude of the improvement varies with the number of layers using DropConnect. We expected that the performance in terms of uncertainty quality and accuracy would be the best with a fully DropConnect network, but our results show that using less DropConnect layers performs best, with minimal differences in the quality of uncertainty. We make the following contributions:

\begin{itemize}
    \item We identify that there is a significant computational speedup in applying MCDC to only a select number of layers with marginal loss in uncertainty quality. We call this method Select-DC.
    \item We find that the best model performance in terms of accuracy happens with partial usage of DropConnect across the network, not with a full DropConnect one, which we believe is unexpected.
    \item In contrast to our expectation, we find that networks with DropConnect applied to select number of layers do not observe significant changes in uncertainty estimation quality. 
    \item We characterize the trade-off between the number of layers MCDC is applied to, and model performance.
    \item We find that DropConnect can be enabled at inference on select layers from the network's output with minimal loss in accuracy and uncertainty quality, enabling dynamic use of DropConnect without network retraining.
\end{itemize}

\subsection{Related Work}

Neural networks with their large number of parameters render the task of predicting the posterior distribution intractable. There has been extensive research on Bayesian neural networks \cite{mackay1992practical} \cite{neal2012bayesian}
and Monte-Carlo sampling for uncertainty estimation in deep learning. While exact Bayesian inference is intractable due to computational costs and challenging inference, several studies have been conducted on approximate methods using deterministic approaches.
\cite{gal2016dropout}
developed a theoretical framework and demonstrated the mathematical equivalence of Dropout training in an arbitrary neutral network with approximate Bayesian inference in deep Gaussian processes \cite{damianou2013deep}.
The prediction ensemble is generated by keeping drop-out at test time. Similar approximations can be done using DropConnect\cite{mobiny2019dropconnect}.
They also introduce an adaptive approach to model the irreducible noise using held-out validation. This proposed a scalable alternative to mean-field variational inference methods, such as Radial BNNs \cite{farquhar2020radial}
and Bayes by Back-prop\cite{blundell2015weight}.
While these class of methods that use Monte-Carlo(MC) sampling work well with estimating the multi-modal distribution, they cannot represent data uncertainty. While a solution that fine-tunes dropout rates has been proposed \cite{gal2017concrete}, \cite{osband2016risk} discusses examples where this method fails to generate correct predictions.

Deep ensembles \cite{lakshminarayanan2017simple} is another frequentist approach towards modeling uncertainty by training multiple models with different random initializations, where each model's parameters could be interpreted as a sample from the underlying weight distribution. While this outperforms Bayesian methods trained using variational inference, it is computationally intensive, and the computation cost, at both train and test time, scales linearly with the number of ensembles. An alternative to this method is \cite{devries2018learning} that learns confidence estimates on the out-of-distribution detection task, without requiring labels for supervised training. The model architecture and loss function formulation is similar in implementation to uncertainty estimation for regression tasks as in \cite{kendall2017uncertainties}, and \cite{gurevich2017learning}. 
There have also been attempts at the above task of out-of-distribution detection using generative models \cite{nalisnick2018deep}, but are computationally expensive than classification models and do not perform predictive uncertainty estimation.

\section{Uncertainty Estimation using DropConnect}
In this section, we briefly review DropConnect and Dropout. We explain how applying DropConnect to all layers approximates a Bayesian NN, and is used to model epistemic uncertainty (the measure of what the model does not know). We then show that we can gain significant computational speedup in estimating uncertainty by applying DropConnect to a select few layers in the model, which we call \textbf{Select-DC}

\subsection{DropConnect and Dropout}
Consider a neural network with $N$ layers. For simplicity, we assume this is a network with simple feed-forward layers. However, it can be easily extended to modern networks like ResNet\cite{he2016deep} and
VGG \cite{simonyan2014very}. We denote the network parameters by $\theta$, and the weight kernel of the layer by $W$. The $o^{th}$ column in the weight matrix denotes the $o^{th}$ neuron in the layer. We ignore bias for convenience, however, biases are not masked in our implementation.

DropConnect \cite{wan2013regularization} randomly drops out individual weights in the weight matrices at every training step. The dropping is actually done by masking weights, i.e, zeroing out their activations. This forces the network to not over-rely on a specific connectivity pattern and adapt to various connectivity patterns. Let $\sigma$ be an activation function. The outputs of a layer will then be 
$$
Y = \sigma (X ( W \odot M )) 
$$
Where $\odot$ is the Hadamard product, $X$ the activations of the previous layer or the input vector if its the first layer, $M$ is the binary mask that randomly drops out weights whose elements are  $\sim$ Bernoulli (p). DropConnect is a generalization of dropping out entire neurons in a network, as in Dropout \cite{gal2016dropout}. Dropout follows the same procedure as above, but instead samples a mask that randomly masks out entire columns of the weight matrix. Since a neuron is completely removed, Dropout is better written as 
$$
Y = \sigma ((X * W) \odot M ) 
$$
The activations of the corresponding neuron are zeroed out. DropConnect at inference has been shown to approximate the Bayesian predictive posterior distribution \cite{mobiny2019dropconnect}, similar to what Dropout \cite{gal2016dropout} does at inference. These methods are called MC-Dropout and MC-DropConnect.

\subsection{Why DropConnect on Select Layers ?}\label{Why DropConnect on select layers ?}

Let us consider the total time complexity of running inference through a trained network, where we apply DropConnect to each convolutional or dense layer. If the network has $N$ layers, and a forward pass through each layer takes approximately $M$ units of time, the total cost of one forward pass is $N * M$. If we run the sample through the network $K$ times, the total time to calculate the statistics of the predictive distribution is $(K * N * M)$. 

State-of-the-art neural networks are usually hundreds, of layers deep. Running multiple forward passes through these networks might be ill-suited for applications that demand low latency and have limited compute. If we apply DropConnect to a select $L$ number of layers, the total time taken to compute $K$ samples would be, 
\begin{equation}
T = (N - L)M + (L * M * K) \leq (K * N * M)
\label{time_eqn}
\end{equation}
The advantage here is that we can vary the layers we apply DropConnect to during inference. We can also train our models using DropConnect applied to all layers(MCDC), and use Select-DC for inference. This flexibility implies that we are not limited to Select-DC if the uncertainty quality is not good enough. We can immediately shift to MCDC to gain the best performance.

Select-DC has one hyper-parameter $\lambda$, which is the number of layers without masked weights during inference. We term this subset of layers as the frozen block.
The frozen block must begin from the input layer to satisfy the equation \ref{time_eqn}. For example, if $\lambda = 4$ in a ResNet20 model, DropConnect is not applied to layers 1-4, and the layers 1-4 form the frozen block. Instead, if we did not apply DropConnect to layers 1, 4, 8, 15 for instance, we cannot store the intermediate activations to reuse them to infer multiple samples. 

Dropping off weights from layers is an approximation of sampling from the underlying weight distribution. We expected that reducing the number of layers DropConnect is applied to, might reduce the uncertainty estimation quality compared to MCDC. Intuitively, if we apply DropConnect to all layers starting from layer x, then every layer before it is common to all forward passes used to estimate the predictive distribution- making these samples less diverse. However, to our surprise, we noticed that there is little to no loss in uncertainty estimation quality using Select-DC.

\section{Experiments}
We evaluate our proposed method on three datasets for image classification: CIFAR10, CIFAR100 and
SVHN. For all the datasets, we use ResNet-20 \cite{he2016deep} with random shifts and horizontal flips as data augmentation. We also use VGG19 \cite{simonyan2014very} to further evaluate the performance of our method on CIFAR10. We report accuracy relative to a baseline model without uncertainty quantification, and all metrics are computed over 25 forward passes of each model. 

 \begin{figure}
     \centering
     \begin{subfigure}[b]{0.4\textwidth}
         \centering
         \includegraphics[width=\textwidth]{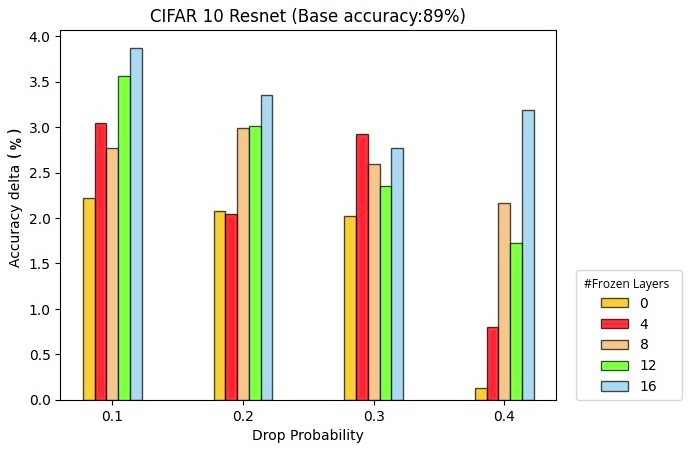}
         \caption{ResNet 20 - CIFAR 10}
         \label{fig:drop prob layer cifar10}
     \end{subfigure}
     \hfill
     \begin{subfigure}[b]{0.4\textwidth}
         \centering
         \includegraphics[width=\textwidth]{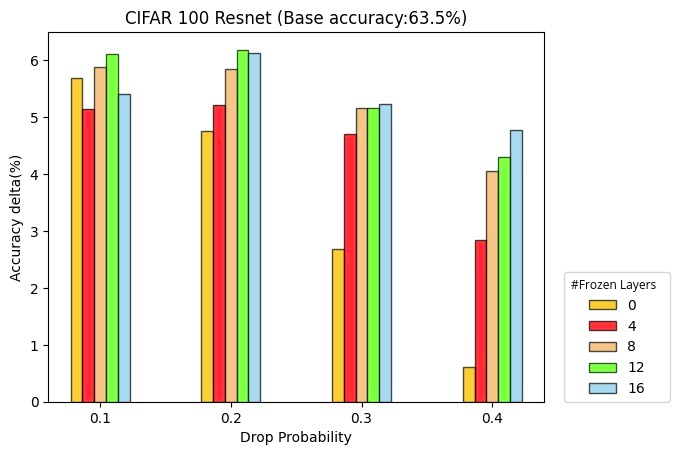}
         \caption{ResNet 20 -CIFAR 100}
         \label{fig:drop prob layer cifar100}
     \end{subfigure}
     
     \begin{subfigure}[b]{0.4\textwidth}
         \centering
         \includegraphics[width=\textwidth]{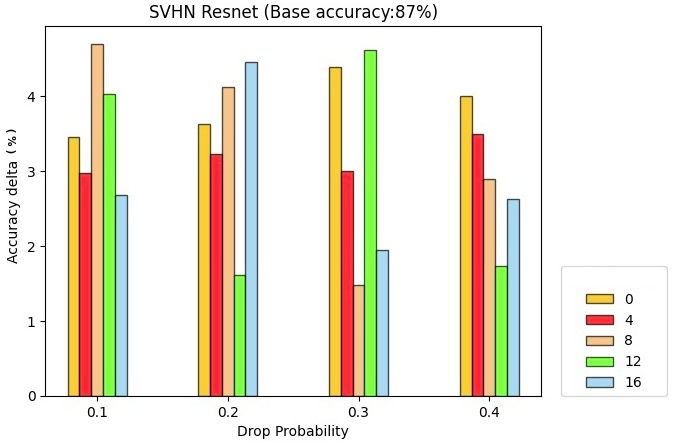}
         \caption{ResNet 20 - SVHN}
         \label{fig:drop prob layer svhn}
     \end{subfigure}
     \hfill
     \begin{subfigure}[b]{0.4\textwidth}
         \centering
         \includegraphics[width=\textwidth]{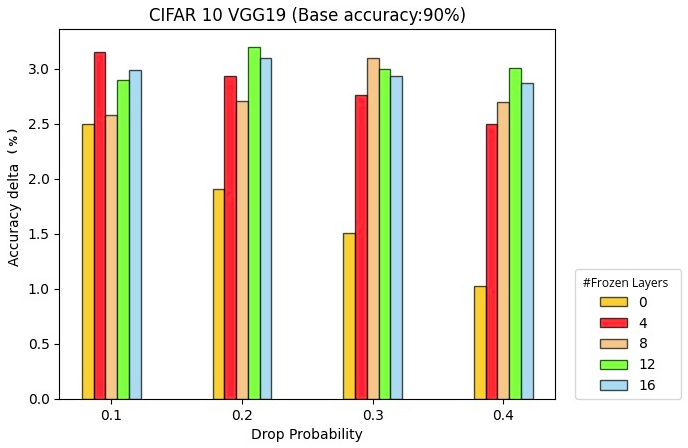}
         \caption{VGG 19 - CIFAR 10}
         \label{fig:drop prob layer vgg cifar10}
     \end{subfigure}
     
        \caption{Comparison of ResNet20 and VGG19 performance on CIFAR10, CIFAR100, SVHN for varying $\lambda$ and drop probabilities.}
        \label{fig:resnet_changing drop prob layer}
    \end{figure}
    
\subsection{Hardware and Setup}
We ran all our experiments on a single machine with an NVIDIA P100 GPU. For both CIFAR 10 and CIFAR 100, all models were trained for 200 epochs with a batch size of 100, and SGD with Nesterov momentum as the optimizer. We varied the learning rate over the course of training, and we maintain a learning rate of 0.5 till 50\% of train time. We then linearly decrease the learning rate from 0.5 to 0.0005 till 90\% of train time. For the last 10\%, we maintain the learning rate at 0.0005. We experimented with various learning rate schedules like linear, exponential decay and cyclical learning rates, but the described schedule performed best. All values reported (accuracy, NLL, entropy) are the mean of 25 samples (stochastic forward passes).
\subsubsection{Select-DC on Training and Inference}

In this experiment, we train the models with $\lambda \geq 0$, and perform inference with the same setting. Fig. \ref{fig:resnet_changing drop prob layer} illustrates the effect of changing $\lambda$ on accuracy. A summary of our experiments on various datasets for combinations of different drop probability and $\lambda$ shows that applying DropConnect to all layers is the least performing setting. This is as expected, since dropout is being applied to all weights in the network. As we decrease $\lambda$, we notice a steady improvement in accuracy. 
   

\subsubsection{MCDC on Training, Select-DC on Inference}

\begin{figure}
    \centering
    \begin{subfigure}[b]{0.40\textwidth}
        \centering
        \includegraphics[width=\textwidth]{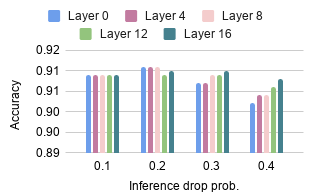}
        \caption{Trained using MCDC with drop prob. 0.1}
        \label{fig:0_1 accuracy}
    \end{subfigure}
    \hfill
    \begin{subfigure}[b]{0.40\textwidth}
        \centering
        \includegraphics[width=\textwidth]{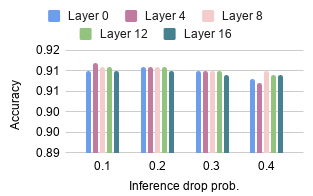}
        \caption{Trained using MCDC with drop prob. 0.2}
        \label{fig:0_2 accuracy}
    \end{subfigure}
    \begin{subfigure}[b]{0.40\textwidth}
        \centering
        \includegraphics[width=\textwidth]{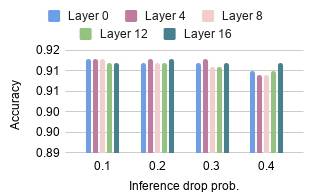}
        \caption{Trained using MCDC with drop prob. 0.3}
        \label{fig:0_3 accuracy}
    \end{subfigure}
    \hfill
    \begin{subfigure}[b]{0.40\textwidth}
        \centering
        \includegraphics[width=\textwidth]{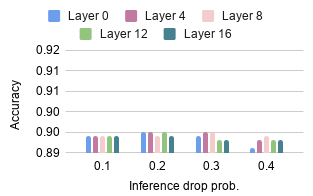}
        \caption{Trained using MCDC with drop prob. 0.4}
        \label{fig:0_4 accuracy}
    \end{subfigure}
    
    \caption{Accuracy Results of training the model using MCDC and inference using Select-DC for varying $\lambda$ and drop probabilities on CIFAR-10. }
    \label{fig:accuracy plots}
\end{figure}

\begin{figure}
    \centering
    \begin{subfigure}[b]{0.40\textwidth}
        \centering
        \includegraphics[width=\textwidth]{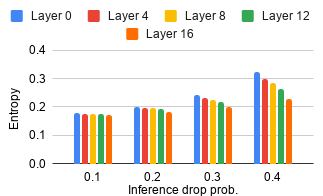}
        \caption{Drop Connect on all layers with drop prob. 0.1}
        \label{fig:0_1 entropy}
    \end{subfigure}
    \hfill
    \begin{subfigure}[b]{0.40\textwidth}
        \centering
        \includegraphics[width=\textwidth]{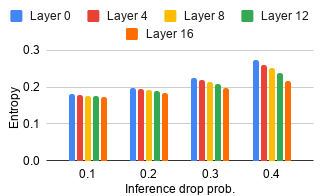}
        \caption{Drop Connect on all layers with drop prob. 0.2}
        \label{fig:0_2 entropy}
    \end{subfigure}
    \begin{subfigure}[b]{0.40\textwidth}
        \centering
        \includegraphics[width=\textwidth]{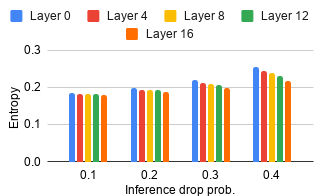}
        \caption{Drop Connect on all layers with drop prob. 0.3}
        \label{fig:0_3 entropy}
    \end{subfigure}
    \hfill
    \begin{subfigure}[b]{0.40\textwidth}
        \centering
        \includegraphics[width=\textwidth]{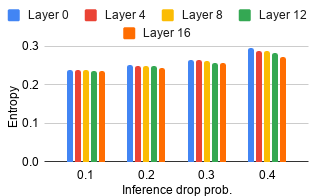}
        \caption{Drop Connect on all layers with drop prob. 0.4}
        \label{fig:0_4 entropy}
    \end{subfigure}
    
    \caption{Results of training the model using MCDC and inference using Select-DC for varying $\lambda$ and drop probabilities on CIFAR-10}
    \label{fig:entropy plots}
\end{figure}

\begin{figure}
    \centering
    \begin{subfigure}[b]{0.32\textwidth}
        \centering
        \includegraphics[width=\textwidth]{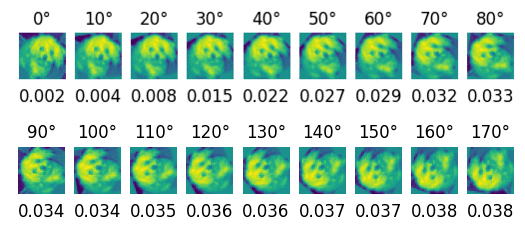}
        \caption{Drop prob 0.1, $\lambda = 0$}
        \label{fig:0 rotate}
    \end{subfigure}
    \hfill
    \begin{subfigure}[b]{0.32\textwidth}
        \centering
        \includegraphics[width=\textwidth]{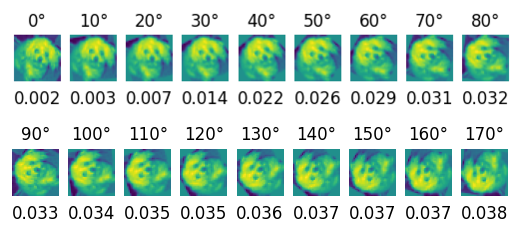}
        \caption{Drop prob 0.1, $\lambda = 8$}
        \label{fig:8 rotate}
    \end{subfigure}
    \hfill
    \begin{subfigure}[b]{0.32\textwidth}
        \centering
        \includegraphics[width=\textwidth]{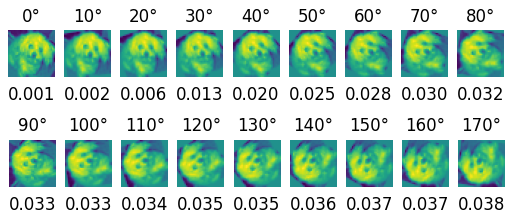}
        \caption{Drop prob 0.1, $\lambda = 16$}
        \label{fig:16 rotate}
    \end{subfigure}
    \caption{Uncertainty estimation for rotated images in CIFAR10. We apply Dropconnect to all layers while training and use Select-DC for inference.}
    \label{fig:rotation}
\end{figure}

In this experiment, we train all of our models with $\lambda = 0$ but perform inference with varying values of $\lambda$. We observe no significant difference in the accuracy of the network on test data. Fig. \ref{fig:entropy plots} illustrates the uncertainty qualities of a model trained with MCDC, and Select-DC used at inference time across different drop probabilities, while Fig. \ref{fig:accuracy plots} shows the corresponding accuracies. Fig. \ref{fig:entropy plots} shows that at lower drop probabilities, the entropy results are almost the same for all values of $\lambda$. As the drop probability at inference time increases, entropy decreases with increasing $\lambda$. These trends are consistent across varying drop probabilities applied during training. Fig. \ref{fig:rotation} further shows how the network's uncertainty changes as we apply a rotational transformation to images from the CIFAR10 dataset.

\subsection{Computational Performance Analysis}
\begin{wrapfigure}{r}{0.4\textwidth}
    \centering
    \vspace*{-1em}
    \includegraphics[scale=0.38]{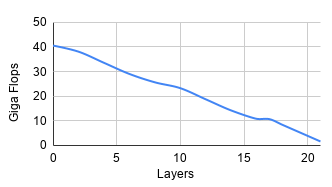}
    \caption{GFLOPS required for 25 forward passes for ResNet20 on CIFAR-10 versus variations of $\lambda$}
    \label{layers_vs_flops}
\end{wrapfigure}

Applying DC on less layers has a theoretical advantage over applying them on the full model. In this section we aim to evaluate this hypothesis and measure the speedup that can be obtained by using DC on select layers instead of the whole network. We estimate the number of floating point operations (FLOPS) as we vary $\lambda$. Fig. \ref{layers_vs_flops} shows how the total number of GFLOPS change with varying $\lambda$, and is consistent with the theory discussed in section \ref{Why DropConnect on select layers ?}. Increasing $\lambda$ decreases the number of GFLOPS required for computation of a forward pass.

\begin{figure}
\centering
     \begin{subfigure}[b]{0.32\textwidth}
         \centering
         \includegraphics[width=\textwidth]{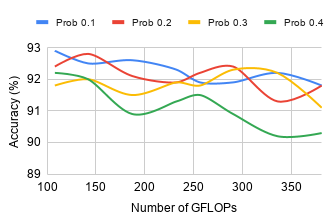}
         \caption{Accuracy vs Giga Flops}
         \label{fig:accuracy vs flops}
     \end{subfigure}
     \hfill
     \begin{subfigure}[b]{0.32\textwidth}
         \centering
         \includegraphics[width=\textwidth]{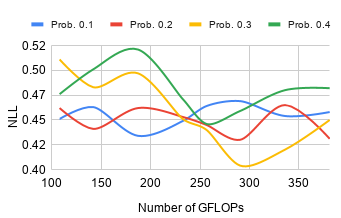}
         \caption{NLL vs Giga Flops}
         \label{fig:nll vs flops}
     \end{subfigure}
     \hfill
     \begin{subfigure}[b]{0.32\textwidth}
         \centering
         \includegraphics[width=\textwidth]{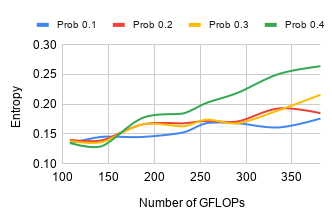}
         \caption{Entropy vs Giga Flops}
         \label{fig:entropy vs flops}
     \end{subfigure}
        \caption{Metrics measured against GFLOPS required to estimate the predictive distribution using 25 samples. The horizontal-axis in the figures correspond to the number of GFLOPS for increasing $\lambda$ values. The GFLOPS decrease as $\lambda$ increases.}
        \label{fig:computational performance}
\end{figure}

We plot the accuracy, negative log-likelihood and entropy as a function of GFLOPS for values of $\lambda \in (0, 21)$ to demonstrate the trade-off between error, uncertainty quality and computational requirements. In Fig. \ref{fig:accuracy vs flops}, we observe that the accuracy decreases with an increase in GFLOPS, i.e, decreasing values of $\lambda$. This is consistent with the trends observed in Fig. \ref{fig:resnet_changing drop prob layer}. In Fig. \ref{fig:entropy vs flops}, we see that the uncertainty of the network, characterized by entropy of the predictive distribution, increases as we decrease $\lambda$, i.e, decreasing number of GFLOPS. However, the loss in quality of uncertainty modeling falls far slower than the required GFLOPS. Particularly for lower drop probabilities, this loss in quality is negligible. Therefore, according to the situational demands, one can convert Select-DC to MCDC to better model epistemic uncertainty.

Our results from Fig. \ref{fig:computational performance} show that while using Select-DC, increasing $\lambda$ has the effect of decreasing the amount of compute (in GFLOPS). At the same time, it reduces the accuracy slightly, around $1-2\%$, depending on the drop probability. This is consistent with other results, where the network with less Bayesian capabilities works as an approximation of the full Bayesian network. It results in lower accuracy and NLL, and slightly increased entropy due to the loss in accuracy producing increased uncertainty.

           
\subsection{Out-Of-Distribution Detection}

We also show the Out-Of-Distribution (OOD) detection capabilities of our model. We train a model on the CIFAR10 dataset, with Select-DC, and evaluate on the SVHN test set for OOD samples. The image sizes are common in these datasets and they have no classes in common. 

To classify a sample as in-distribution or OOD, we calculate the entropy of predictive distribution estimated through Monte Carlo sampling. The entropy is defined as, 
$$
H(x) = -\sum_{c \in C}f(x)_{c}logf(x)_{c}
$$
We then classify all samples which result in entropy higher than a threshold as OOD, and in-distribution otherwise. An OOD input causes the network to output an approximately uniform distribution, which is observed from the high entropy values. Table \ref{tab:ood_cifar10_svhn} shows our quantitative results with different drop probabilities and $\lambda$ values. Our results show that Select-DC can detect OOD samples nearly as well, if not exactly, as the models with full MCDC, with a difference of less than $1 \%$ AUC across different values of $\lambda$.

\begin{table}[]
    \begin{tabular}{ p{3cm}p{2cm}p{2cm}p{2cm}p{2cm} }
    \toprule
    \multicolumn{5}{c}{Entropy of predictive distribution} \\
    \midrule
    Drop probability& $\lambda$  & ID Entropy & OOD Entropy & AUC\\
    \midrule
        & 0 &0.177&   0.747		& 0.883\\
        & 4  & 0.161  & 0.765	& 0.890\\
  0.1 	& 8 & 0.168&  0.769		& 0.883\\
        & 12 & 0.145&  0.759	& 0.893\\
        & 16 & 0.134&  0.764	& 0.896\\
    \midrule
        & 0 &0.196&   0.731		& 0.887\\
        & 4  & 0.193   & 0.738	& 0.877\\
  0.2 	&  8 & 0.172& 0.729 	& 0.885\\
        & 12 & 0.166&  0.723	& 0.878\\
        & 16 & 0.14&  0.722		& 0.887\\
    \midrule
        & 0 &0.219&   0.732		& 0.866\\
        & 4  & 0.19 & 0.730		& 0.868\\
 0.3  	& 8 & 0.174&  0.706		& 0.874\\
        & 12 & 0.166 &  0.732	& 0.871\\
        & 16 & 0.138 &  0.695	& 0.888\\
    \bottomrule
  \end{tabular}
    \caption{Quantitative Out of Distribution results between CIFAR10 (ID) and SVHN (OOD).}
    \label{tab:ood_cifar10_svhn}
\end{table}

\section{Conclusions and Future Work}
In this work, we present the idea of applying DropConnect only to a select few layers instead of all layers in a neural network to model epistemic uncertainty. We show that we can achieve significant computational speedup by running the intermediate activations through the DropConnect applied part of the network without significant trade-off to uncertainty estimation quality. We show that this can also be used, without remarkable loss in performance, for Out-of-Distribution detection. We present and discuss how changing the subset of layers DropConnect is applied to affects the accuracy, NLL, entropy of a neural network. We experiment on CIFAR 10, CIFAR 100, and SVHN with ResNet and VGG models. We are excited to see how these observations extend to multiple domains like Natural Language Processing, or Reinforcement Learning.

Some limitations of SelectDC are the requirement that the frozen block must be at the beginning of the network, the unexpected loss of performance when the whole network uses DropConnect, which we believe requires further research, and we also expected larger differences in out of distribution detection performance, which might indicate that MC-DropConnect does not produce good epistemic uncertainty quantification.

\clearpage
\printbibliography

\clearpage
\appendix

\section{Broader Impact}
In security critical applications like autonomous driving, the perceptions models are usually trained on well-curated datasets, for example, with very good lighting and environment conditions. Even balancing the dataset by collecting samples of different conditions cannot cover all possible situations that the network might encounter. Here, uncertainty estimation and OOD detection is a pivotal requirement. However, a naive implementation of Bayesian NNs or even approximation techniques are computationally demanding. Our proposed method reduces computational requirements for uncertainty modeling and can be altered according to the requirements. For example, in the perception module of a self driving car, we can apply MCDC only to a few layers. During situations that are easy to interpret like broad day light, run the activations of the last frozen layer through the network to estimate samples of the predictive distribution. During night times, or times where the input to the perception module is noisy, we can apply MCDC to all layers in the network to get the best possible uncertainty estimates.

\section{DropConnect as Bayesian approximation}

Similar to Mobiny et al. \cite{mobiny2019dropconnect}, in this section we prove that DropConnect approximates a Bayesian Neural Network. We provide the proof here for completeness. In a Bayesian NN with $N$ layers, with weights $W = \{W\}_{i=1}^{N}$, our task is calculate the posterior distribution of the weights, $p=(W|D)$, given a dataset $D = (x, y)$. The predictive distribution of a label $y'$ given a sample $x'$ is, 
$$
p(y' | x', D) = \mathbb{E}_{p(w|D)} [p(y'|x', w)] 
$$
$$
= \int p(y' | x', w)p(w|D)dw
$$
However, evaluating the integral for all weights in the weight space is clearly computationally intractable, and neither can it be evaluated analytically. Intuitively, this is equivalent to estimating the predictive distribution an infinite number of times, each time with a different weight configuration, and ensembling them to make a prediction. 
One way to approximate the posterior on the weights is to use variational inference. We use a variational distribution on the weights, $q_{\theta}(w)$ parameterized by $\theta$, to minimize the Kullback-Leibler divergence between $q$ and the true posterior. This is equivalent to minimizing negative evidence lower bound and takes the form,
\begin{equation} \label{loss}
    L(\theta) = - \int q_{\theta}(w) log(p(y|x, w))dw + KL(q_{\theta}(w)||p(w))
\end{equation}

We can develop an accurate approximation, generalizing the approach followed in \cite{gal2016dropout}, using Monte Carlo sampling. 

We approximate the variational distribution $q(w_{k}|\theta_{k})$ for layer $k$ as $w_{k} = \theta_{k} \odot M_{k}$, where $M_{k}$ is the binary mask sampled from a Bernoulli distribution and $\theta_{i}$, the variational parameters to be optimized. 

Rewriting the first term in \ref{loss} as sum over all samples in the dataset,
\begin{equation}
    - \int q_{\theta}(w) log(p(y|x, w))dw = \sum_{n=1}^{N} q_{\theta}(w) log(p(y|x, w)) = \frac{1}{N} \sum_{n=1}^{N} q_{\theta}(w) log(p(y|x, w'))
\end{equation}
Applying DropConnect to weights can be interpreted as $w'$, which is a sample from the weight distribution. The second term in \ref{loss} can be approximated using $\sum_{i=1}^{L} ||\theta||_{2}^{2}$ as shown in \cite{gal2016dropout}.

The loss function then finally takes the form,
\begin{equation}
    \hat{L_{mc}} =  \frac{1}{N} \sum_{n=1}^{N} q_{\theta}(w) log(p(y|x, w')) + \lambda \sum_{i=1}^{L} ||\theta||_{2}^{2}
\end{equation}
During inference, we can replace the posterior $p(w|D)$ with the approximate posterior $q_{\theta}(w)$ and use Monte Carlo sampling to approximate the integral. 
\begin{equation}
    p(y' | x', D) \approx \frac{1}{T}\sum_{t=1}^{T} p(y' | x', w'_{t})
\end{equation}
Each forward pass through the network generates a Monte Carlo sample from the posterior. 
\end{document}